\documentclass[letterpaper, 10 pt, journal, twoside]{IEEEtran}
\usepackage{graphicx}
\usepackage{amsmath}
\usepackage{amssymb}
\usepackage{booktabs}
\usepackage{color}
\usepackage{multirow}

\usepackage{soul}
\usepackage{url}
\usepackage{hyperref}

\usepackage{verbatim}
\usepackage{subfigure}
\usepackage{calc}
\usepackage{amstext}

\usepackage{stackengine} 
\newcommand\oast{\stackMath\mathbin{\stackinset{c}{0ex}{c}{0ex}{\ast}{\bigcirc}}}

\newcommand\IncG[2][]{\addstackgap{%
\raisebox{-.5\height}{\includegraphics[#1]{#2}}}}
\usepackage{array}
\usepackage[capitalize]{cleveref}
\crefname{section}{Sec.}{Secs.}
\Crefname{section}{Section}{Sections}
\Crefname{table}{Table}{Tables}
\crefname{table}{Tab.}{Tabs.}

\ifCLASSINFOpdf
\else
\fi
\begin{document}

% \title{Explicit Attention-Enhanced RGB-Thermal Fusion for Perception Tasks}
\title{Explicit Attention-Enhanced Fusion for RGB-Thermal Perception Tasks}

\author{Mingjian Liang$^{*}$, Junjie Hu$^{*}$, Chenyu Bao, Hua Feng, Fuqin Deng and Tin Lun Lam$^{\dagger} $% <-this % stops a space
% \thanks{*This work was not supported by any organization}% <-this % stops a space
% \thanks{This paper is partially supported by funding 2019-INT008 from the Shenzhen Institute of Artificial Intelligence and Robotics for Society.}% <-this % stops a space
\thanks{J. Hu, H. Feng, F. Deng, T.L. Lam are with the Shenzhen Institute of Artificial Intelligence and Robotics for Society.}%
\thanks{M. Liang, C. Bao and T.L. Lam are with the School of Science and Engineering, The Chinese University of Hong Kong, Shenzhen.}%
\thanks{$*$ indicates equal contribution.}
\thanks{$^{\dagger}$Corresponding author: Tin Lun Lam
        {\tt\small tllam@cuhk.edu.cn}
        }%
      \thanks{  This work was partly supported by the National Natural Science Foundation of China (62073274), Shenzhen Science and Technology Program (JCYJ20220818103000001), and
 the funding AC01202101103 from the Shenzhen Institute of Artificial Intelligence and Robotics for Society.}
}



\maketitle

\begin{abstract}

    Recently, RGB-Thermal based perception has shown significant advances. Thermal information provides useful clues when visual cameras suffer from poor lighting conditions, such as low light and fog. However, how to effectively fuse RGB images and thermal data remains an open challenge. Previous works involve naive fusion strategies such as merging them at the input, concatenating multi-modality features inside models, or applying attention to each data modality. 
    These fusion strategies are straightforward yet insufficient. 
    In this paper, we propose a novel fusion method named Explicit Attention-Enhanced Fusion (EAEF) that fully takes advantage of each type of data. Specifically, we consider the following cases: i) both RGB data and thermal data, ii) only one of the types of data, and iii) none of them generate discriminative features. EAEF uses one branch to enhance feature extraction for i) and iii) and the other branch to remedy insufficient representations for ii).
    The outputs of two branches are fused to form complementary features. As a result, the proposed fusion method outperforms state-of-the-art by 1.6\% in mIoU on semantic segmentation, 3.1\% in MAE on salient object detection, 2.3\% in mAP on object detection, and 8.1\% in MAE on crowd counting. The code is available at https://github.com/FreeformRobotics/EAEFNet.

\end{abstract}

% \begin{IEEEkeywords}
% IEEE, IEEEtran, journal, \LaTeX, paper, template.
% \end{IEEEkeywords}
\begin{IEEEkeywords}
Multi-modality data fusion, RGB-Thermal fusion, RGB-thermal perception
\end{IEEEkeywords}

% \fi
\IEEEpeerreviewmaketitle

\section{Introduction}
% {\color{blue}
Over the last decade, we have witnessed significant progress on many perception tasks. Based on data-driven learning, deep neural networks (DNNs) can learn to estimate semantic maps \cite{FCN}, object categories \cite{Yolo}, depth maps \cite{Hu2019RevisitingSI}, etc., from only RGB images. This paradigm has continuously boosted perception tasks for robots in which various models, loss functions, and learning strategies have been explored.

However, current methods highly depend on the quality of RGB images. In reality, visual cameras are particularly susceptible to noises \cite{suganuma2019attention}, poor lighting \cite{hu2021two}, weather \cite{liu2019dual}, etc. In these cases, DNNs tend to degrade their performance significantly. To handle these issues, researchers sought to employ thermal data to complement RGB images and developed different multi-modality fusion strategies.

The core of RGB-T based methods is the fusion strategy of RGB data and thermal data. Previous methods \cite{tardal,cddfuse} directly combine them at the input. Some works \cite{MFNet,RTFNet} use two separate encoders for extracting features from RGB and thermal images, respectively. Then, these features are merged and outputted to a decoder to yield a final prediction.  
 Recently, most studies \cite{ABMDRNet,FEANet} attempted to utilize the attention mechanism for multi-modality data fusion. These approaches commonly apply channel attention to intermediate features of different data types and obtain the fused features by weighing their importance. However, these fusion strategies are implicit and insufficient. In particular, it is unclear how multi-modality data can (or cannot) complement each other.
 % {\color{red} How Add, CWF and FEAM fuse RGB and thermal data? }

\begin{figure}[t!]
    \centering
    \includegraphics [width=0.48\textwidth]{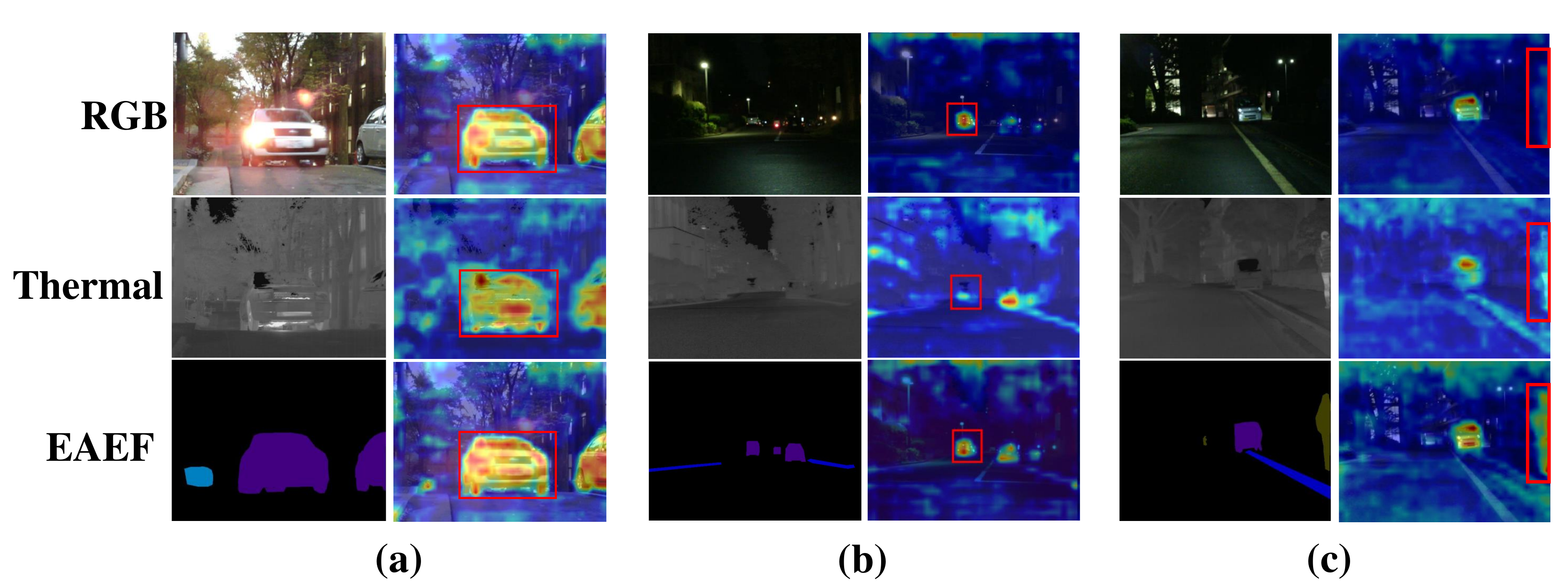}
    \vspace{-3mm}
    \caption{Visualization of extracted features of RGB, thermal images, and the proposed fusion method, where (a) both RGB and thermal data, (b) only one of them, and (c) none, can yield distinct features. As seen, the proposed fusion method can boost feature extraction for all three cases.}
    \vspace{-3mm}
    % Illustration of modality difference reduction. ((a),(b),(c)) represent the cars and pedestrians that need to be identified in the original RGB , Thermal and Ground Truth, ((d),(g)) grad cam\cite{grad-cam} after ResNet\cite{ResNet} feature extraction, ((e),(f),(h),(i)) respectively grad cam after Add\cite{RTFNet}, FEAM\cite{FEANet}, CWF\cite{ABMDRNet}, IDAM}
    \label{fig_intro}
\end{figure}
% {\color{red}
Different from existing studies, we explicitly take multi-modality data fusion under three circumstances: i) both RGB and thermal images can extract useful features, as Fig.~\ref{fig_intro} (a), ii) only one of them can generate meaningful representations, as Fig.~\ref{fig_intro} (b), and iii) none of them provides useful features, as Fig.~\ref{fig_intro} (c). 
In this paper, we propose the Explicit Attention-Enhanced Fusion (EAEF) that performs a more effective fusion.
The key inspiration of EAEF is the case-specific design that uses one branch to stick to meaningful representations for i) and enhance feature extraction for iii), and the other branch to force CNNs to pay attention to insufficient representations for ii). One of these branches will generate useful features at least, and their combination will yield complementary features for a final prediction.

 \begin{figure*}[t!]
    \centering
    \includegraphics [width=0.96\textwidth]{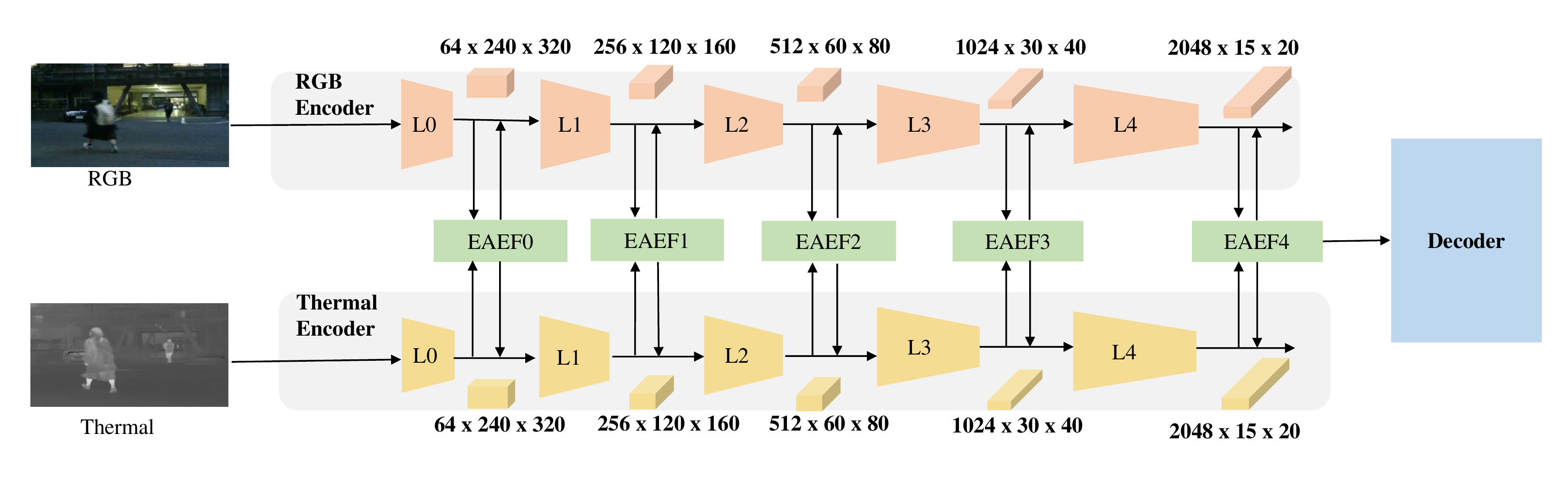}
    \vspace{-5mm}
    \caption{Diagram of encoder-decoder based network assembled with the proposed EAEF for dense prediction tasks. EAEF is used for fusing features from the RGB branch and the thermal branch at multi-scales. }
    \label{fig_network}
    \vspace{-3mm}
\end{figure*}

To validate the effectiveness of this novel multi-modality fusion method, we design a novel RGB-T framework by integrating EAEF into an encoder-decoder network and evaluate it on various vision tasks with benchmark datasets, including semantic segmentation, object detection, and crowd counting. We confirm through extensive experiments that our method is more effective for RGB-T based vision perception.

In summary, our contributions are:
\begin{itemize}
    \item A novel multi-modality fusion method that effectively fuses RGB features and thermal features in an explicit manner.
    \item An effective encoder-decoder based network assembled with the proposed feature fusion strategy for dense prediction tasks.
    \item State-of-the-art performance on semantic segmentation, object detection, salient detection, and crowd counting with open-source codes.
\end{itemize}
% }

The remainder of this paper is organized as follows. Section II reviews related studies. Section III presents the framework and the proposed multi-modality data fusion method. Section IV provides quantitative and qualitative experimental results on three tasks. Section V concludes our work.

%-------------------------------------------------------------------------
%-------------------------------------------------------------------------
\section{Related Work}
\label{sec_related_work}

Using additional modality data to complement RGB images has shown great improvement in accuracy. Many works attempted to fuse depth and RGB data using attention mechanisms \cite{liu2020learning,Hu2022DeepDC,zhou2020rgb} . 
For example, FRNet \cite{FRNet} and HFNet\cite{HFNet} apply channel attention for RGB-D fusion. BCINet \cite{BCINet} uses spatial attention in their BCIM (Bilateral cross-modal interaction module) to capture cross-modal complementary features. RLLNet \cite{rllnet} leverages both channel and spatial mechanisms in the decoder. Similarly, methods of RGB-T perception have explored the fusion strategy between RGB data and thermal data. In this paper, we mainly review related works from the perspective of feature fusion.
\paragraph{RGB-T semantic segmentation}

Early works fuse RGB features and thermal features in a straightforward way by applying element-wise addition, such as MFNet \cite{MFNet}, RTFNet \cite{RTFNet}, FuseSeg \cite{FuseSeg}, neglecting the difference of features in their importance on discriminability. Recent works solve this issue by employing channel attention as in ABMDRNet \cite{ABMDRNet}, both channel and spatial attention as in FEANet \cite{FEANet} and GMNet \cite{GMNet}, self-attention as in MFTNet \cite{MFTNet}. Besides, the domain adaptation from day to night using thermal images has also been studied in \cite{heatnet,MS-UDA}.

\paragraph{RGB-T Salient Objection detection}

Attention mechanism has also been frequently used in the salient detection task, such as the multi-interaction module using channel attention proposed in MIDD \cite{MDDI}.
In most cases, both channel and spatial attention are employed to weigh channels and pixels adaptively, e.g., cross-guidance fusion in
SCGFNet \cite{CGFNet}, cross-modal attention mechanism in \cite{FMCF}, Convolutional Block Attention Module (CBAM) in ADFNet \cite{ADFNet}. Similarly, based on attentional considerations, LSNet \cite{lSNet} introduced attention selection to determine inter-attention at each pixel position.

\paragraph{RGB-T crowd counting}

% In 2021, IADM recognized the potential for combining thermal cameras used for detecting human body temperature during the COVID-19 pandemic with RGB cameras to perform crowd counting tasks in complex lighting conditions. To support the development of RGBT, they also introduced a new benchmark dataset called RGBT-CC. Building on previous research on multi-modal perception of RGB-T, they proposed an IADM module that provides a complementary information exchange approach to facilitate information sharing between modalities.

% DEFNet\cite{} is built on the DenseNet architecture, allowing for effective extraction of fused features from both RGB and thermal modalities. Through the use of convolutional layers, it can effectively suppress background interference. Furthermore, by fusing features from different layers, it can capture contextual positional information for objects of similar sizes, thereby improving recognition accuracy.

% CSCA \cite{CSCA} proposed an attention module that concurrently considers feature information and feature spatial information to address the issue of spatio-temporal misalignment between modalities in the crowd counting task, effectively mitigating accuracy errors. Subsequently, TAFNet\cite{} was proposed based on the IADM module, which added attention mechanism to enhance the extraction of complementary information between modalities and further improved the accuracy of information exchange.

DEFNet \cite{DEFNet} fuses the two modalities through simple element-wise addition. IADM \cite{iadm} and TAFNet \cite{TAFNet} introduce a gate mechanism for feature fusion. Not surprisingly, an attempt to incorporate attention mechanism has also been carried out by CSCA \cite{CSCA}.
% where self-attention is transferred from unimodality to multi-modality by generating the query from another modality. Also, as we argued above, this simple modification of self-attention doesn't explicitly take the contribution from each modality into account.

\paragraph{RGB-T objection detection}
Existing works including TarDAL \cite{tardal}, CDDFuse \cite{cddfuse}, and U2F\cite{u2f}, directly combine RGB data and thermal data at inputs, without applying attention based fusion strategy. 
% This early fusion strategy generally yields worse discriminability, but is good at reducing model complexity.

However, there is no guarantee that existing methods utilizing attention can extract sufficiently effective features since feature generation and attention extraction are performed implicitly inside CNNs. In this paper, we propose a case-specific way to improve the classical channel attention mechanism by explicitly enhancing attention extraction.

\begin{figure*}[t!]
    \centering
    \includegraphics[width=0.9\textwidth]{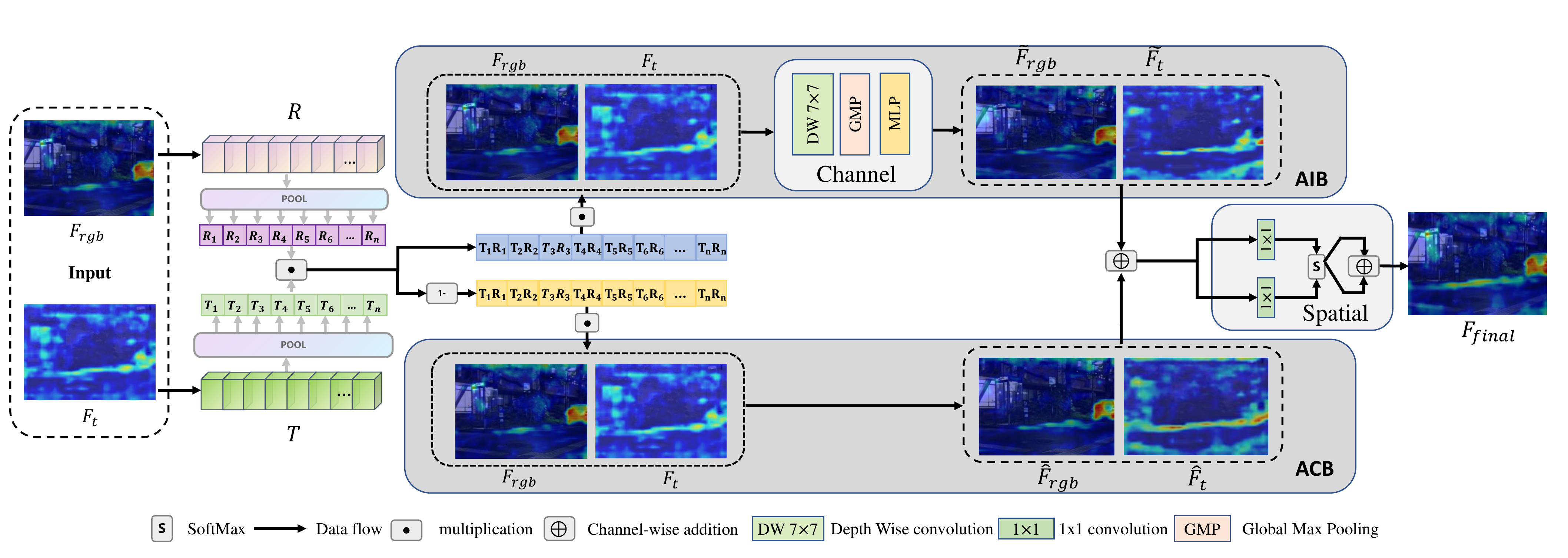}
    \vspace{-2mm}
    \caption{Overview of the proposed Explicit Attention-Enhanced Fusion (EAEF). EAEF takes RGB features $F_{rgb}$ from the RGB encoder and thermal features $F_t$ from the thermal encoder as inputs, then applies attention interaction and attention complement with two branches. The final features are fused by merging outputs from these two branches. For simplicity, we only show one feature map extracted from the image encoder and the thermal encoder, respectively.
     \vspace{-3mm}
    % The inputs are an RGB steam feature and T steam feature.$\text { Then } \text { number }_i=\sum_{u, v} \mathbb{1}\left[\arg \max _c s_{u, v}^{(c)}=i\right]$ \cite{PKD}
    }
    \label{fig_eaef}
    % \vspace{-5mm}
\end{figure*}

%------------------------------------------------------------------------
\section{Methodology}
\label{sec:intro}

 \subsection{Framework Overview}
 We take the classical encoder-decoder network for dense prediction tasks. The framework consists of an image encoder, a thermal encoder, and a decoder.
 % Similar to existing approaches, the RGB encoder is used to extract features from RGB images and the thermal encoder is used to extract features from thermal images. 
 The proposed Explicit Attention-Enhanced Fusion (EAEF) is applied between the two encoders to fuse features at multi-scales. A diagram of our dense prediction network is given in Fig.~\ref{fig_network}, where we show a semantic segmentation network built on ResNet \cite{ResNet}. 
 % The framework uses five convolutional blocks to extract multi-scale features; thus, we apply five EAEF modules to fuse RGB and thermal features. 
 
 Note that the framework naturally uses different backbones on different tasks. Therefore, the detailed implementation is task-specific. Nevertheless, all models built on our framework have the same technical components, i.e., one RGB encoder, one thermal encoder, EAEF modules, and a decoder.
 % In experiments, we will use this framework to evaluate the effectiveness of our method on three different vision tasks.

\subsection{Explicit Attention-Enhanced Fusion}

% \subsubsection{Channel-wise Attention Extraction}

%------------------------------------------------------------------------

% \subsubsection{Connect RGB and T matrix}
Suppose $F_{rgb}\in\mathbb{R}^{h \times w \times c}$ and $F_{t}\in\mathbb{R}^{h \times{w} \times{c}}$ are the features extracted from RGB encoder and thermal encoder at a certain scale, 
the conventional way to extract channel-wise attention can be represented as
% we first quantify whether $F_{rgb}$ or $F_{t}$ contains sufficiently discriminative features.
% extract their channel-wise attention. Following the common strategy \cite{clip},
% We apply the global average pooling to $F_{rgb}$ and $F_{t}$ along the channel dimension and then apply an MLP to obtain the weights as follows:
\begin{equation}
 \begin{aligned}
& R = \biggl( f_{MLP} \Bigl( f_{GAP}(F_{rgb}) \Bigr) \biggr) \\
& T = \biggl( f_{MLP} \Bigl( f_{GAP}(F_{t}) \Bigr) \biggr)
 \end{aligned}
 \label{eq_at}
\end{equation}
where $R \in\mathbb{R}^{c}$ and $T \in\mathbb{R}^{c}$ are extracted weights for RGB features and thermal features, respectively;
$f_{GAP}$ and $f_{MLP}$ denote the global average pooling and MLP, respectively. 
For many previous works, the feature fusion is conducted by 
$\sigma(R) \oast F_{rgb} + \sigma(T) \oast F_{t}$, where $\sigma$ is the sigmoid activation that generates channel-wise attention, $\oast$ denotes channel-wise multiplication. However, this fusion method is effective only if either $\sigma(R)$ or $\sigma(T)$ has been activated sufficiently.

Differing from any existing approaches, we delve into this feature fusion by explicitly considering the interaction of multi-modality features. 
% We explicitly specify four cases for extracted attention.
% as seen in Table~\ref{attention_cases} where $R$ and $T$ are extracted weights of feature maps by Eq.~\eqref{eq_at}. Noting that $R$ and $T$ are vectors, we treat them as scalars in Table~\ref{attention_cases} for simplicity. The positive values denote higher attention, i.e., $\sigma(R)\in[0.5,1)$; similarly, the negative values denote lower attention, i.e., $\sigma(R)\in(0, 0.5]$. 
% For all of these cases, we apply attention enhancement to generate higher attention for both RGB and thermal features.
Specifically, we decompose the feature fusion into an Attention Interaction Branch (AIB) and an Attention Complement Branch (ACB), as shown in Fig.~\ref{fig_eaef}. The former handles cases where both RGB and thermal encoder or none of them capture discriminative features, and the latter tackles cases where only one encoder extracts useful features.

% \subsubsection{Attention Interaction Branch (AIB)}
AIB takes an element-wise multiplication between $R$ and $T$ to generate correlated attention, then applies channel-wise multiplication to RGB and thermal features. It is represented as:

\begin{equation}
 \begin{aligned}
&  F'_{rgb}  = \sigma( c * (R \otimes T)) \oast F_{rgb}   \\
& F'_{t}  = \sigma( c * (R \otimes T)) \oast F_{t}
 \end{aligned}
 \label{eq_cab}
\end{equation}
where $\otimes$ denotes element-wise multiplication.
$c$ is the number of channels that plays a role of scaling factor for attention enhancement, such that $  \sigma( c * (R \otimes T)) \geq  \sigma(R)$ and $  \sigma( c * (R \otimes T)) \geq \sigma(T)$.% \approx  \sigma(T)$ .}

% \subsubsection{Attention Complement Branch (ACB)}
For cases where only one modality data provides sufficiently discriminative features, i.e., $R\geq0$, $T\leq0$ or $R\leq0$, $T\geq0$, $\sigma( c \oast (R \otimes T))$ tends to be small. Thus, we use the attention complement branch (ACB) that applies the enhancement by:
\begin{equation}
 \begin{aligned}
&  \hat{F}_{rgb}  = (1 - \sigma( c * (R \otimes T))) \oast F_{rgb}  \\
& \hat{F}_{t}  = (1 - \sigma( c * (R \otimes T))) \oast F_{t}
 \end{aligned}
 \label{eq_eab}
\end{equation}
such that $ 1 - \sigma( c * (R \otimes T)) \geq  \sigma(R)$ and $ 1 - \sigma( c * (R \otimes T)) \geq \sigma(T)$. 

 At least one branch can generate stronger attention than the traditional channel-wise attention mechanism. However, if we directly combine $\hat{F}_{rgb}$ and $F'_{rgb}$ or $\hat{F}_{t}$ and $F'_{t}$, it will lead to an identical mapping, e.g., $\hat{F}_{rgb} + F'_{rgb} = F_{rgb}$. To tackle this issue, we apply non-linear feature interaction within each branch.

% It is observed that Eq.\eqref{eq_cab} is most effective if both $R$ and $T$ are large. 
Let $F'_{rgb,t}$ be the concatenation of $F'_{rgb}$ and $F'_{t}$; then, AIB further performs multi-modality interaction between $F'_{rgb}$ and $ F'_{t}$ by a data interaction module:
\begin{equation}
\tilde{F}_{rgb,t} = F'_{rgb,t} \oast \sigma \biggl( f_{MLP} \biggl( f_{GMP} \Bigl( f_{dw}(F'_{rgb,t}) \Bigr)\biggr) \biggr) \\
 \label{eq_cab2}
\end{equation}
where $f_{MLP}$ and $\sigma$ denote MLP and sigmoid operations which are the same as Eq.\eqref{eq_at}, $f_{dw}$ is depth-wise convolution, $f_{GMP}$ is global max pooling operation. $\tilde{F}_{rgb,t}$ is the outputted features by the data interaction module, and it is further split to RGB features $\tilde{F}_{rgb}$ and thermal features $\tilde{F}_{rgb}$, respectively.

We experimentally found that the interaction module contributes less to the ACB regarding the model's performance. Therefore, we do not apply the interaction module in ACB to reduce the model complexity.

Then, the enhanced RGB and thermal features are obtained by aggregating outputs from AIB and ACB:
\begin{equation}
 \begin{aligned}
 & \overline{F}_{rgb}  = \tilde{F}_{rgb} + \hat{F}_{rgb} \\
  & \overline{F}_{t}  = \tilde{F}_{t} + \hat{F}_{t} 
   \end{aligned}
\end{equation}

Finally, we apply a spatial attention mechanism to merge the enhanced RGB and thermal features with a 1$\times$1 convolutional layer. Formally, the merged features are obtained by:
\begin{equation}
  F^{*}_{rgb,t}  = \overline{F}_{rgb,t} \otimes SoftMax\Bigl( Conv_{1\times1} (\overline{F}_{rgb,t} )\Bigr)
\end{equation}
where $\overline{F}_{rgb,t}$ denotes the concatenated result of $\overline{F}_{rgb}$ and $\overline{F}_{t}$. 
The outputted features of EAEF are obtained by:
\begin{equation}
F_{final}=F^{*}_{rgb} + F^{*}_{t}
\end{equation}
 % is the fused features outputted by the EAEF module and is sent to the RGB encoder and the thermal encoder, respectively. 

\section{Experimental Results}
\label{sec_results}

%--------------------------------------------------------------------------------------------------------

%------------------------------------------------------------------------%--------------------------------
\begin{table*}[htbp]
  \centering
  \caption{Quantitative comparisons on the MFNet dataset. The best and the second best results are shown in bold font and the color blue, respectively.}
  \setlength{\tabcolsep}{0.3mm}
    \begin{tabular}{ccccccccccccccccccccc}
    \toprule
    \multicolumn{1}{c}{\multirow{2}{*}{Method}} & \multicolumn{1}{c}{\multirow{2}{*}{Flops}}&\multicolumn{1}{c}{\multirow{2}{*}{Params.(M)}} & \multicolumn{2}{c}{Car} & \multicolumn{2}{c}{Person} & \multicolumn{2}{c}{Bike} & \multicolumn{2}{c}{Curve} & \multicolumn{2}{c}{Car stop} & \multicolumn{2}{c}{Guardrail} & \multicolumn{2}{c}{Color Cone} & \multicolumn{2}{c}{Bump} & \multirow{2}{*}{mAcc} &\multirow{2}{*}{mIoU} \\
\cmidrule{4-19}   &  &  & Acc   & IoU   & Acc   & IoU   & Acc   & IoU   & Acc   & IoU   & Acc   & IoU   & Acc   & IoU   & Acc   & IoU   & Acc   & IoU     \\
\cmidrule{1-21}
\multicolumn{1}{c}{MFNet\cite{MFNet}} &0.74   & 8.4 & 77.2  & 65.9  & 67.0  & 58.9  & 53.9  & 42.9  & 36.2  & 29.9  & 19.1  & 9.9   & 0.1   & 8.5   & 30.3  & 25.2  & 30.0  & 27.7  & 45.1  & 39.7  \\
    \midrule
    \multicolumn{1}{c}{FuseNet(VGG16)\cite{Fusenet}} & 44.17  &284.0  & 81.0  & 75.6  & 75.2  & 66.3  & 64.5  & 51.9  & 51.0  & 37.8  & 17.4  & 15.0  & 0.0   & 0.0   & 31.1  & 21.4  & 51.9  & 45.0  & 52.4  & 45.6  \\
    \midrule
    \multicolumn{1}{c}{PSTNet(ResNet18)\cite{pst900}} &123.4  & 105.8 & -     & 76.8  & -     & 52.6  & -     & 55.3  & -     & 29.6  & -     & 25.1  & -     & \textbf{15.1}  & -     & 39.4  & -     & 45.0  & -     & 48.4  \\
    \midrule
    \multicolumn{1}{c}{RTFNet(ResNet50)\cite{RTFNet}} & 185.2 &245.7 & 91.3  & 86.3  & 78.2  & 67.8  & 71.5  & 58.2  & 59.8  & 43.7  & 32.1  & 24.3  & 13.4   & 3.6   & 40.4  & 26.0  & 73.5  & 57.2  & 62.2  & 51.7  \\
    \midrule
    \multicolumn{1}{c}{ABMDRNet(ResNet50)\cite{ABMDRNet}} & -  &-  & 94.3  & 84.8  & \textbf{90.0}  & 69.6  & 75.7  & 60.3  & 64.0  & 45.1  & 44.1  & 33.1  & 31.0  & 5.1   & 61.7  & 47.4  & 66.2  & 50.0  & 69.5  & 54.8  \\
    \midrule
    \multicolumn{1}{c}{GMNet(ResNet50)\cite{GMNet}} &153.0 & 149.8  & 94.1  & 86.5  & 83.0  & \textbf{73.1} & 76.9  & 61.7  & 59.7  & 44.0  & \textbf{55.0}  & \textbf{42.3}  & \textbf{71.2}  & 14.5   & 54.7  & 48.7  & 73.1  & 47.4  & 74.1  & 57.3  \\
    \midrule
    \multicolumn{1}{c}{FuseSeg(DenseNet161)\cite{FuseSeg}} & -  &-  & 93.1  & \textbf{87.9}  & 81.4  & 71.7  & 78.5  & \textbf{64.6}  & 68.4  & 44.8  & 29.1  & 22.7  & 63.7  & 6.4   & 55.8  & 46.9  & 66.4  & 49.7  & 70.6  & 54.5  \\
    \midrule
    \multicolumn{1}{c}{FEANet(ResNet152)\cite{FEANet}} &255.2 & 337.1  & 93.3  & 87.8  & 82.7  & 71.1  & 76.7  & 61.1  & 65.5  & 46.5  & 26.6  & 22.1  & 70.8& 6.6   & \textbf{66.6} & 55.3  & \textbf{77.3} & 48.9  & 73.2  & 55.3  \\
    \midrule
    \multicolumn{1}{c}{RTFNet(ResNet152)\cite{RTFNet}} & 245.5 &337.1  & 91.3  & 87.4  & 79.3  & 70.3  & 76.8  & 62.7  & 60.7  & 45.3  & 38.5  & 29.8  & 0.0   & 0.0   & 45.5  & 29.1  & 74.7  & 55.7  & 63.1  & 53.2  \\
    \midrule
    \multicolumn{1}{c}{EGFNet(ResNet152)\cite{EGFNet}} &62.8 &201.3 & \textbf{95.8}  & 87.6  & 89.0  & 69.8  &80.6  & 58.8  & \textbf{71.5}  & 42.8  & 48.7  & 33.8  & 33.6  & 7.0   & 65.3  & 48.3  & 71.1  & 47.1  & 72.7  & 54.8  \\
    \midrule
    \multicolumn{1}{c}{MFTNet(ResNet152)\cite{MFTNet}} &330.6 &360.9  & 95.1  & \textbf{87.9} & 85.2  & 66.8  & \textbf{83.9} & 64.4 & 64.3  & 47.1  & {50.8} & 36.1 & 45.9  & 8.4   & 62.8  & \textbf{55.5} & 73.8  & \textbf{62.2} & 74.7  & 57.3  \\
    \midrule
    \multicolumn{1}{c}{Ours(ResNet50)} &77.9 &109.1   & 93.9  & 86.8  & 84.6  & 71.8  & 80.4  & 62.0  & 66.8  & \textbf{49.7} & 43.5  & 29.7  & 58.5  & 7.1   & 61.8  & 50.9  & 70.9  & 46.7  & 73.2 & 55.9 \\
    \multicolumn{1}{c}{Ours(ResNet152)} &147.3 &200.4  & 95.4 &87.6  & 85.2  & 72.6  & 79.9  & 63.8  & 70.6  & 48.6  & 47.9  & 35.0  & 62.8  & 14.2 & 62.7  & 52.4  & 71.9  & 58.3  & \textbf{75.1} & \textbf{58.9} \\
    \bottomrule
    \end{tabular}%
  \label{res_mfnet}%
\end{table*}%

\begin{figure*}[t!]
    \centering
    \includegraphics [width=\textwidth]{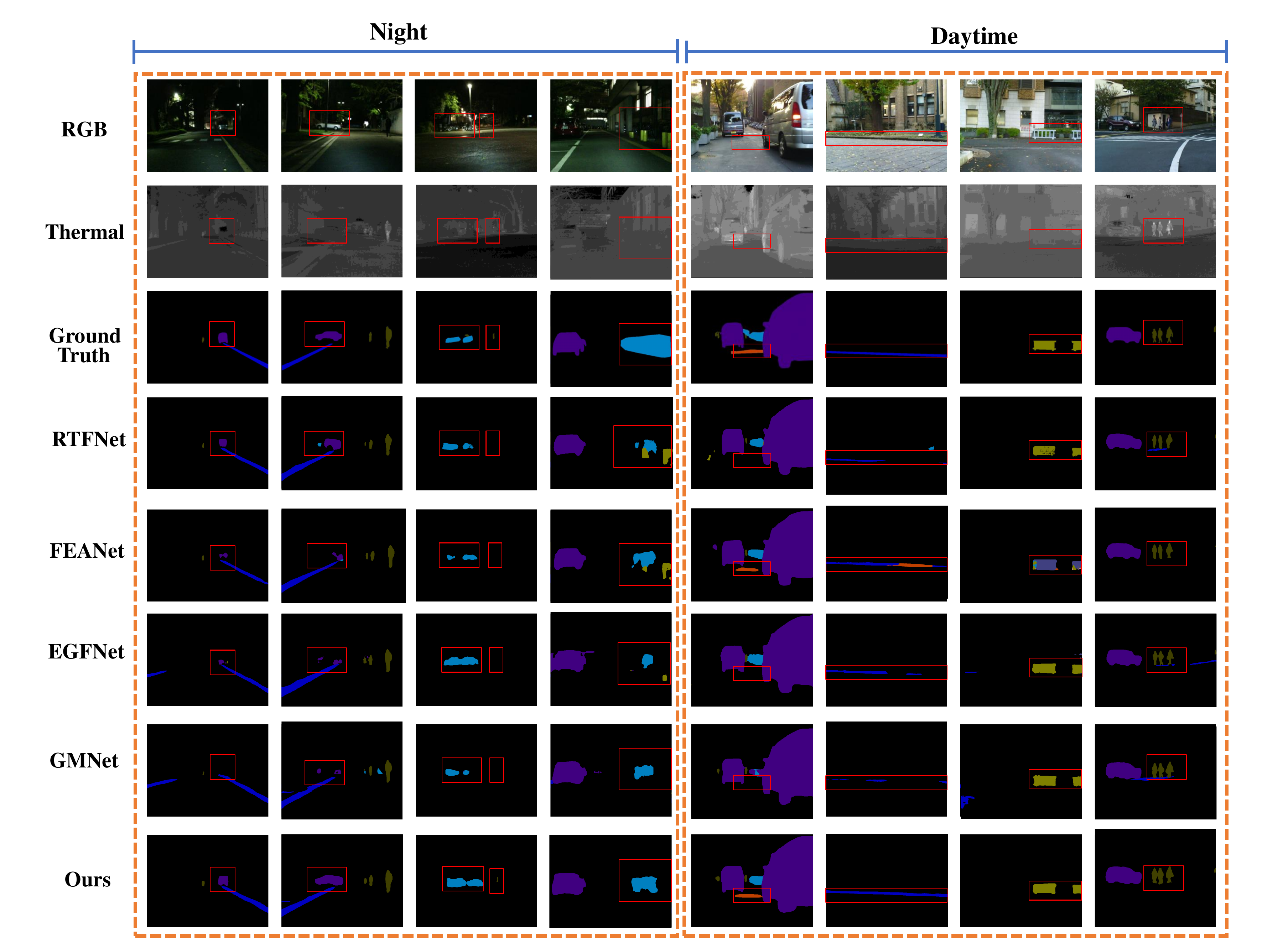}
    % \vspace{-3mm}
    \caption{Qualitative comparisons on the MFNet dataset. We can see that our EAEFNet can provide better results in various lighting conditions and environments. The comparison results demonstrate our superiority.}
    \label{qua_mfnet}
    % \vspace{-5mm}
\end{figure*}

\begin{table}[t!]
  \centering
  \caption{Results on the PST900 dataset.}
  \setlength{\tabcolsep}{0.75mm}
    \begin{tabular}{cccc}
    \toprule
    \multicolumn{2}{c}{Method} & mACC & mIoU \\
    \midrule
    %\multicolumn{2}{c}{Efficient FCN (3C)\cite{EfficientFCN}} & 60.72  & 50.98  \\
    \multicolumn{2}{c}{Efficient FCN\cite{EfficientFCN}} & 66.75  & 57.27  \\
    %\multicolumn{2}{c}{CCNet (3C)\cite{CCNet}}  & 69.71  & 61.42  \\
    \multicolumn{2}{c}{CCNet\cite{CCNet}}  & 73.43  & 66.00  \\
    \multicolumn{2}{c}{ACNet\cite{Acnet}} & 78.67  & 71.81  \\
    \multicolumn{2}{c}{SA-Gate\cite{SAGate}} & 84.71  & 79.05  \\
    \multicolumn{2}{c}{RTFNet\cite{RTFNet}}  & 65.69  & 60.46  \\
    \multicolumn{2}{c}{PSTNet\cite{pst900}} & /     & 68.36  \\
    \multicolumn{2}{c}{GMNet\cite{GMNet}}  & 89.61  & 84.12  \\
    \multicolumn{2}{c}{Ours}  & \textbf{91.42}  & \textbf{85.56} \\
    \bottomrule
    \end{tabular}%
  \label{res_pst}%
\end{table}%

\begin{figure*}[!t]
\centering  
\begin{tabular}
{p{0.085\textwidth}<{\centering}p{0.085\textwidth}
<{\centering}p{0.085\textwidth}
<{\centering}p{0.085\textwidth}
<{\centering}p{0.085\textwidth}
<{\centering}p{0.085\textwidth}<{\centering}p{0.085\textwidth}<{\centering}p{0.085\textwidth}<{\centering}p{0.085\textwidth}<{\centering}} 
% \\
\IncG [width=0.1\textwidth]{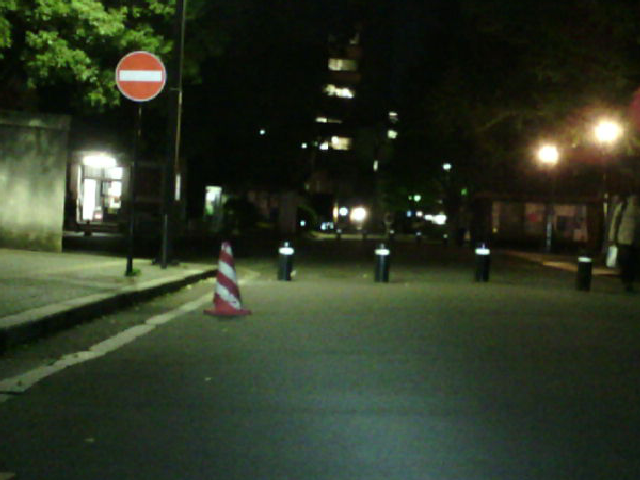} 
&\IncG [width=0.1\textwidth]{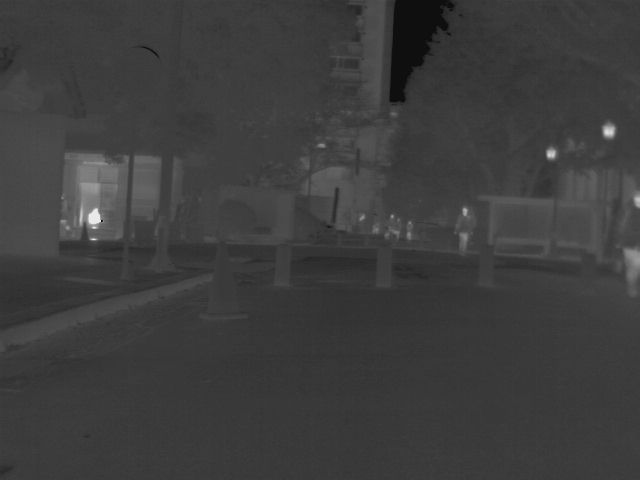} 
&\IncG [width=0.1\textwidth]{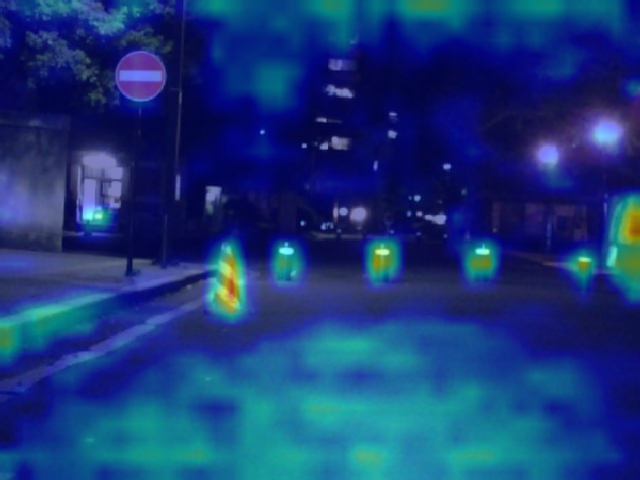} 
&\IncG [width=0.1\textwidth]{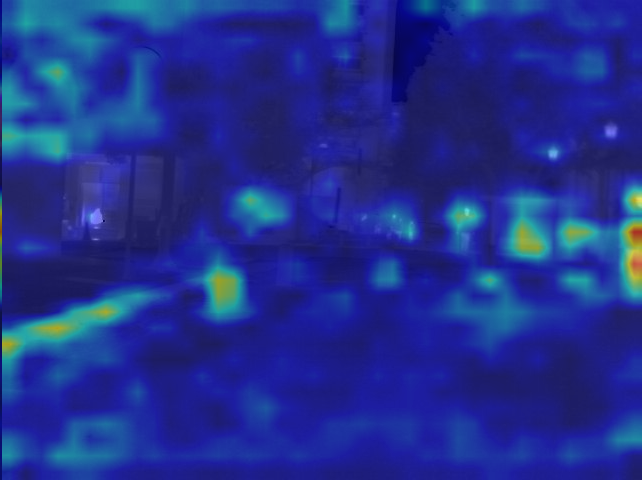}
&\IncG [width=0.1\textwidth]{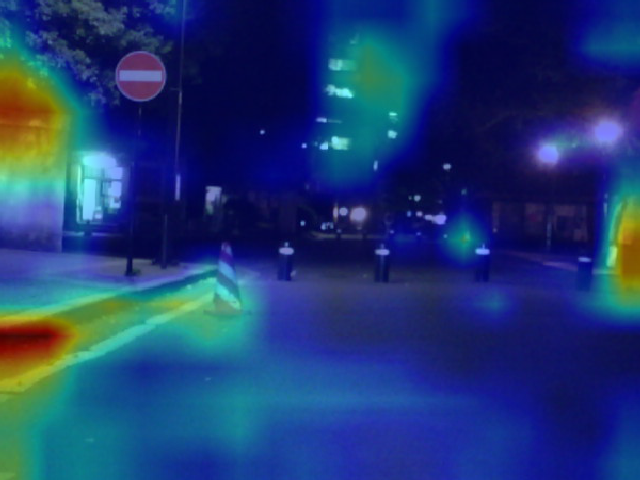} 
&\IncG [width=0.1\textwidth]{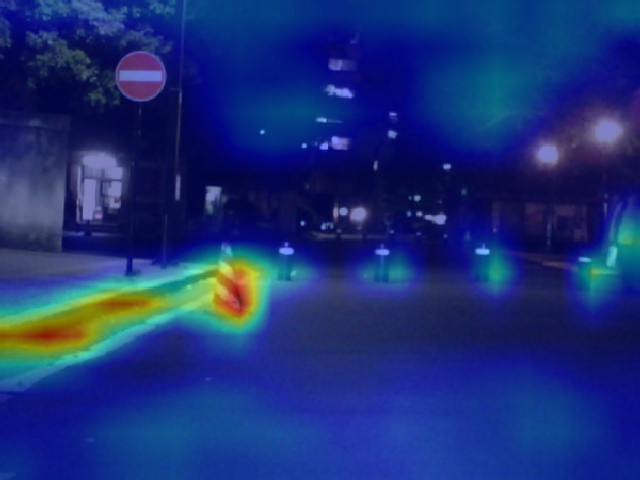} 
&\IncG [width=0.1\textwidth]{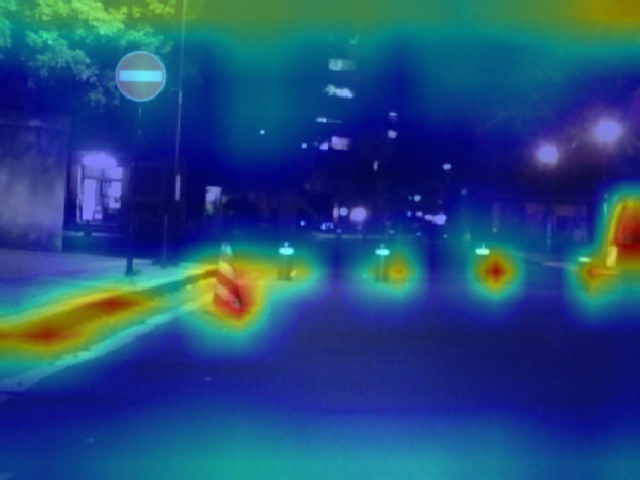}
&\IncG [width=0.1\textwidth]{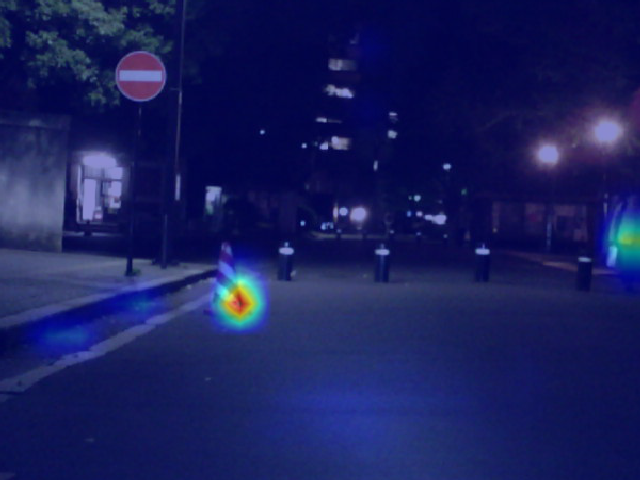}
&\IncG [width=0.1\textwidth]{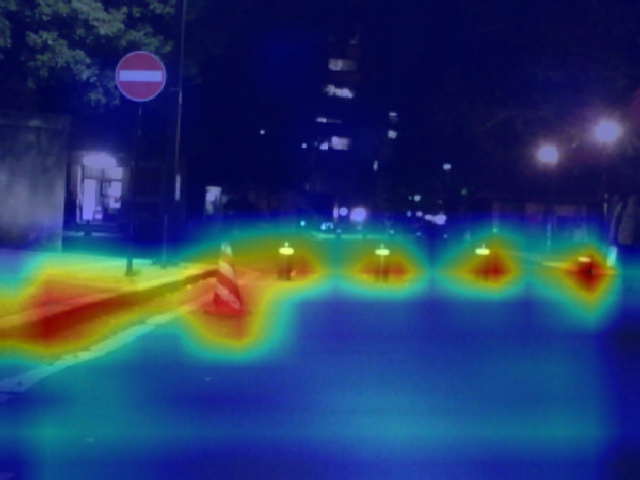}
\\
% \\
\IncG [width=0.1\textwidth]{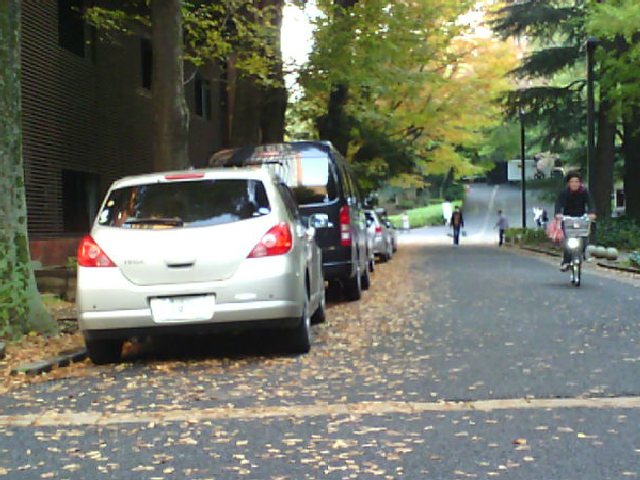}
&\IncG [width=0.1\textwidth]{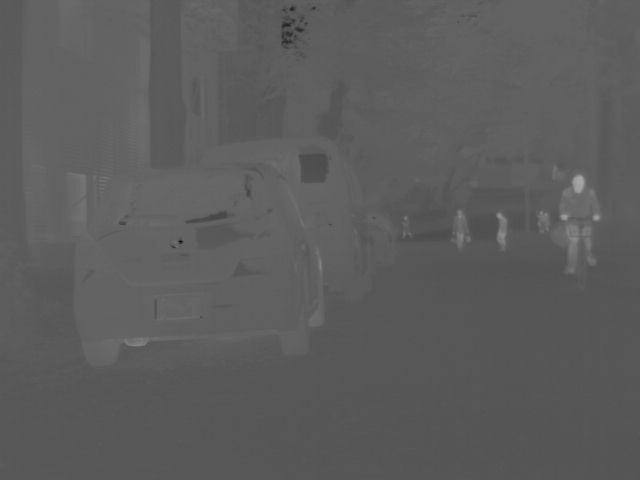} 
&\IncG [width=0.1\textwidth]{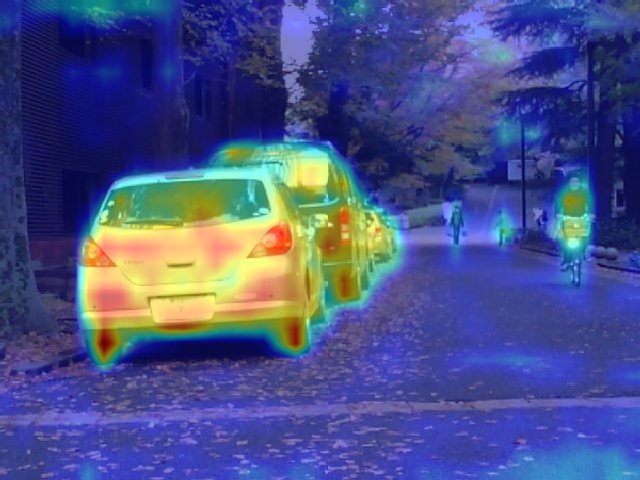} 
&\IncG [width=0.1\textwidth]{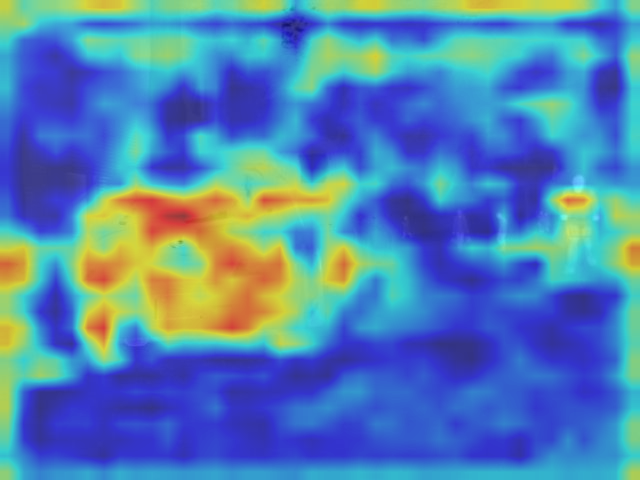}
&\IncG [width=0.1\textwidth]{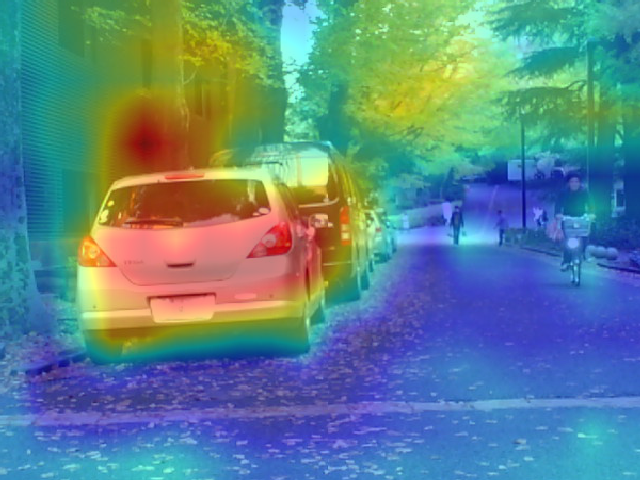} 
&\IncG [width=0.1\textwidth]{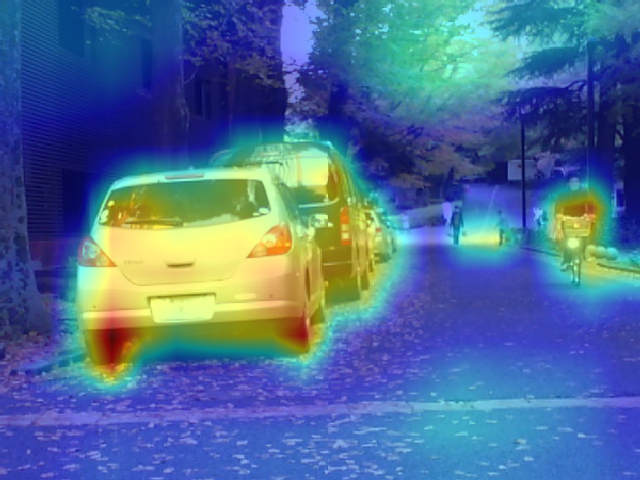} 
&\IncG [width=0.1\textwidth]{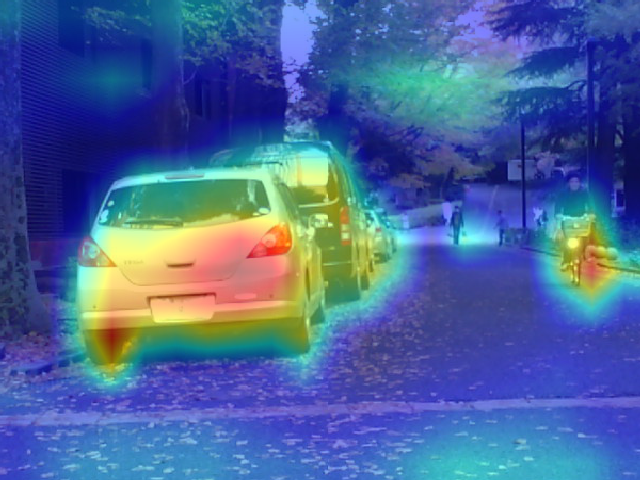}
&\IncG [width=0.1\textwidth]{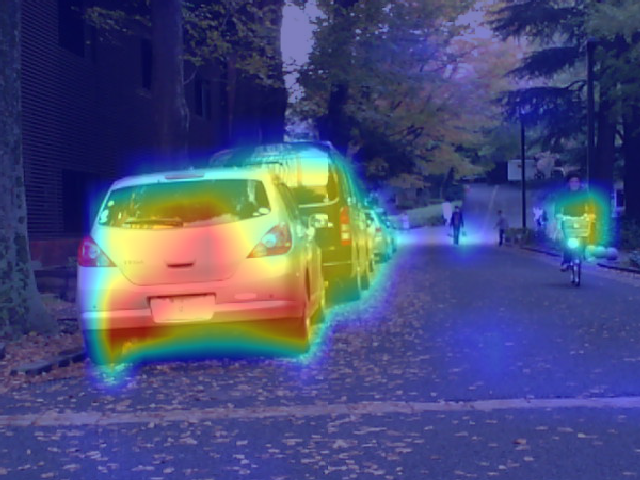}
&\IncG [width=0.1\textwidth]{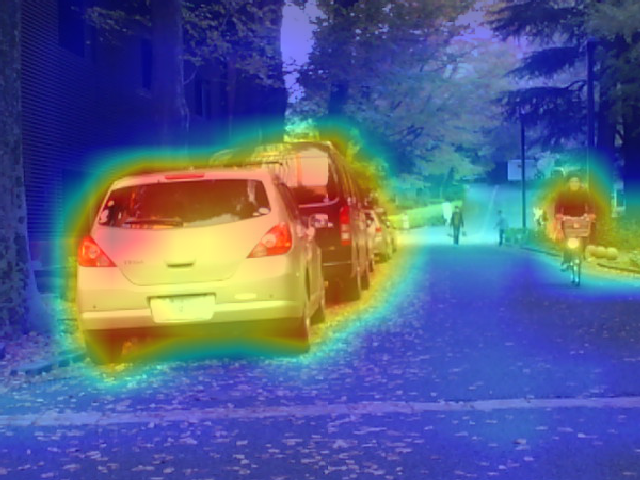}
\\
\IncG [width=0.1\textwidth]{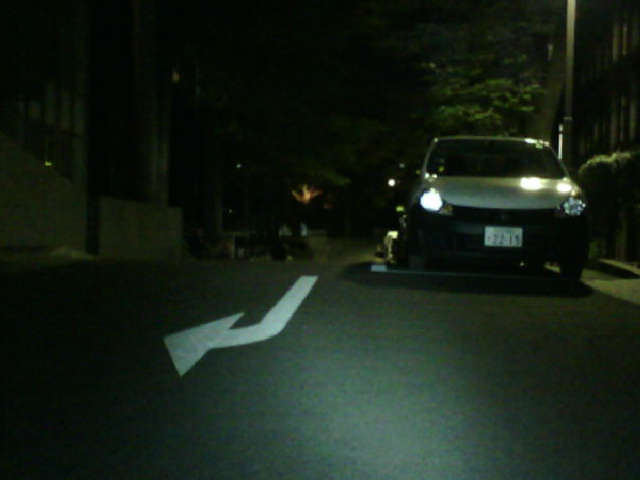} 
&\IncG [width=0.1\textwidth]{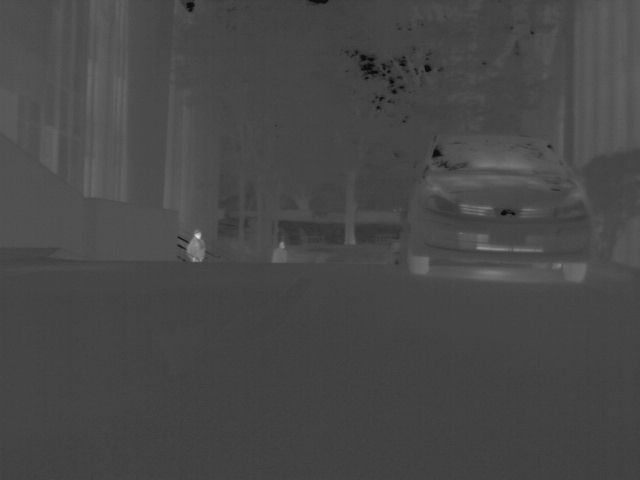} 
&\IncG [width=0.1\textwidth]{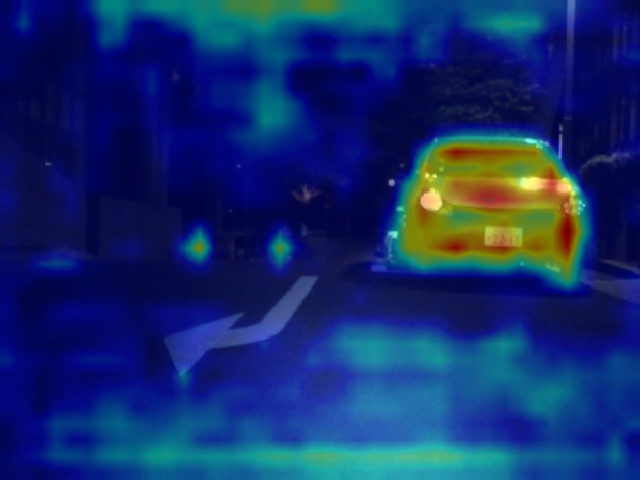} 
&\IncG [width=0.1\textwidth]{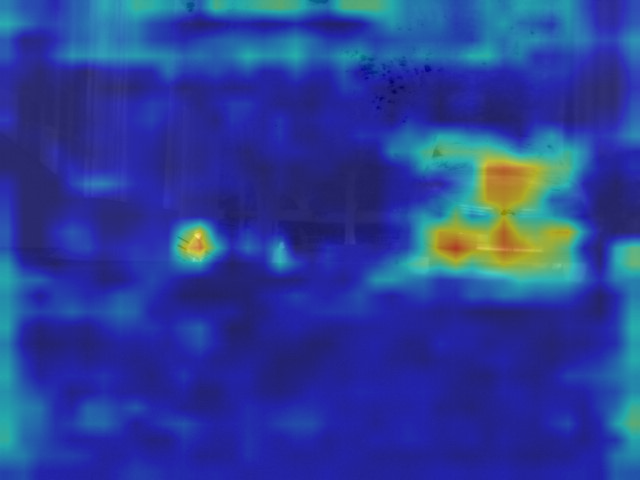}
&\IncG [width=0.1\textwidth]{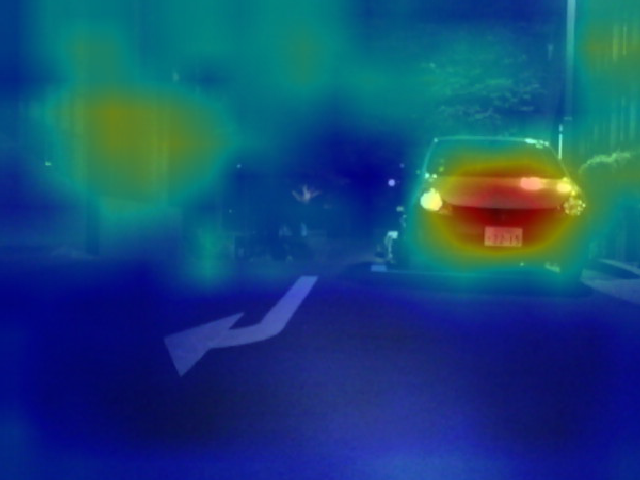} 
&\IncG [width=0.1\textwidth]{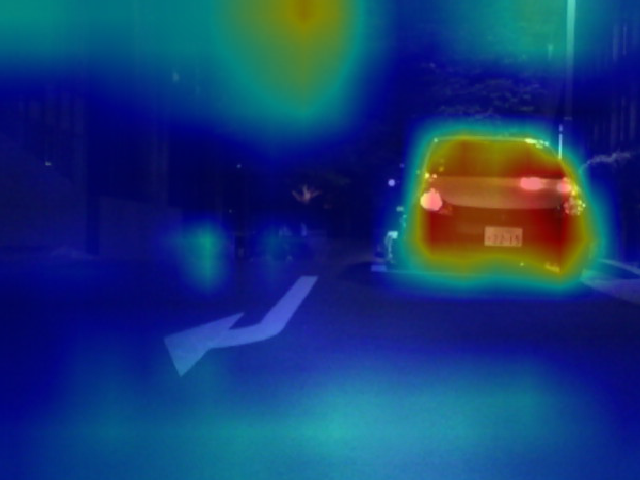} 
&\IncG [width=0.1\textwidth]{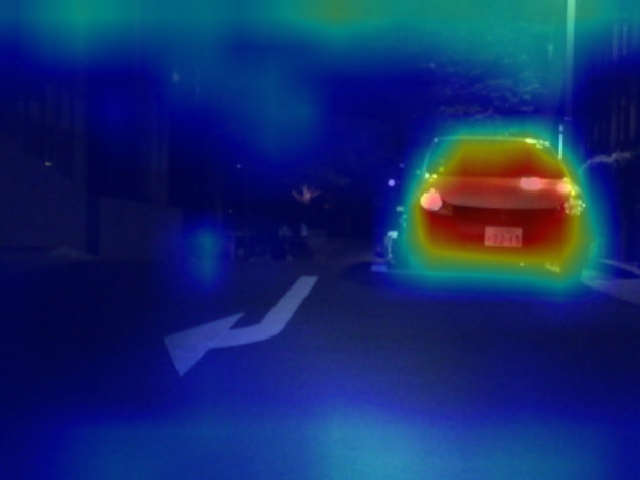}
&\IncG [width=0.1\textwidth]{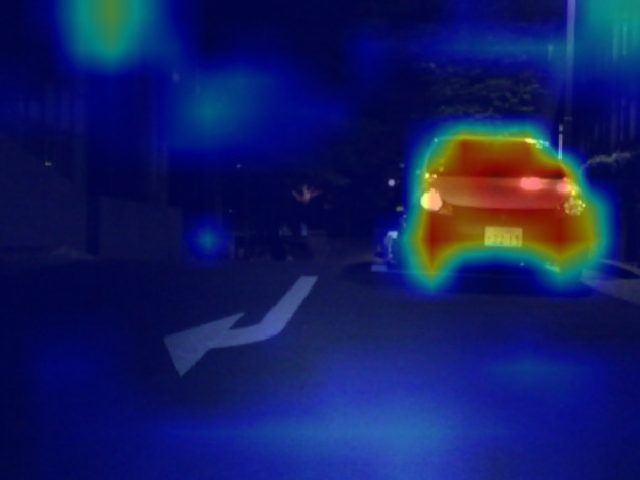}
&\IncG [width=0.1\textwidth]{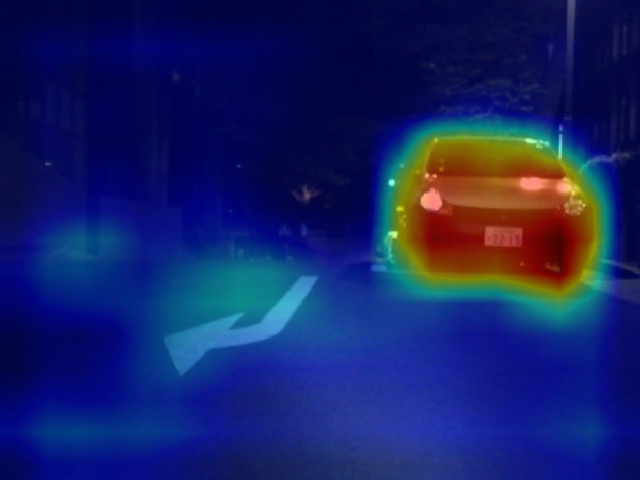}
\\
{\footnotesize RGB}
&{\footnotesize Thermal}
&{\footnotesize RGB attention}
&{\footnotesize Thermal attention}
&{\footnotesize RTFNet \cite{RTFNet}}
&{\footnotesize ABMDRNet\cite{ABMDRNet}}
&{\footnotesize GMNet \cite{GMNet}}
&{\footnotesize FEANet \cite{FEANet}}
&{\footnotesize EAEF}
\\
\end{tabular}
\caption{Visualization of attention maps of different methods on the MFNet dataset.}
 \vspace{-3mm}
\label{Attention Map}
\end{figure*}
% \subsection{Result and Analysis}

\subsection{Semantic Segmentation}

\subsubsection{Datasets}

MFNet dataset \cite{MFNet} is the most popular benchmark for RGB-T based semantic segmentation. It records nine semantic categories in urban street scenes, including one unlabeled background category and eight hand-labeled object categories. The dataset contains 1569 pairs of RGB and thermal images with a resolution of $640 \times 480$. 
Following RTFNet \cite{RTFNet}, we use 784 pairs of images for training, 392 pairs for validation, and the rest 420 pairs for testing.
% Our dataset partition is the same as in RTFNet\cite{RTFNet}, and the training set consists of 784 pairs of images. The validation set consists of 392 teams of images—the other 420 teams' collection of images used for testing.

PST900 dataset \cite{pspnet} is also a popular benchmark for RGB-T based semantic segmentation. It contains five semantic categories and 894 RGB-T image pairs with a resolution of $ 720 \times 1280 $. Among them, 597 pairs are split for training, and the rest 297 pairs are used for testing. 
% There are five semantic categories, including XXXXXX.
% with annotations on 4 semantic categories from the DARPA\cite{DARPA} and other one is background.

\subsubsection{Implementation Details and Evaluation Metrics}
%------------------------------------------------------------------------
% Our experimental environment is set up on an NVIDIA RTX 3090Ti GPU, the network construction and training are performed in Pytorch, and 
We employ a cascaded decoder structure based on BBSNet \cite{BBSNet} with three modifications. First, we replace the asymmetric convolution in GCM with SELayer \cite{SENet} to reduce the model parameters. Second, we introduce an ASPP \cite{DeepLabv3+} structure into Cascaded Decoder to better aggregate low-level features. Finally, we adjust the output dimension of the decoder to meet our nine-categories recognition requirement. We use the stochastic gradient descent (SGD)\cite{SGD} optimization solver for training. The initial learning rate is set to 0.02, Momentum and weight decay are set to 0.9 and 0.0005, respectively.  The batch size is set to 5, and we apply ExponentialLR to gradually decrease the learning rate. 
% Different from previous work, we refer to the solution of the current sample unevenness work, and adopt the method of 
% We employ a mixing loss functions
The loss function has a 
DiceLoss \cite{Dice} term and a SoftCrossEntropy \cite{soft} term, each term is weighed with a scalar of 0.5.
For MFNet dataset, we train the model with 100 epochs and use the best model on the validation set for evaluation.
For PST900 dataset, we train the model with 60 epochs. 

Same to previous works \cite{RTFNet}, we use two measures for quantifying results.
 The first is Accuracy (Acc) and the second one is Intersection Union (IoU). mAcc and mIoU are the averages over all categories.

\subsubsection{Results}
\paragraph{Results on MFNet}
We first conduct quantitative comparisons between the proposed method and other baseline approaches. We compare our method against existing approaches, including MFNet \cite{MFNet}, FuseNet \cite{Fusenet}, RTFNet-152 \cite{RTFNet}, FusSeg-161 \cite{FuseSeg}, FEANet \cite{FEANet}, GMNet \cite{GMNet}, MFTNet \cite{MFTNet}, PSTNet \cite{pst900}, RTFNet-50 \cite{RTFNet}, ABMDENet \cite{ABMDRNet}, EGFNet \cite{EGFNet}. Since the model complexity is different for existing methods, we implement our method on two backbones, including a larger ResNet-152 and a smaller ResNet-50, for fair comparisons. 

Table~\ref{res_mfnet} shows the quantitative results. It is clear that our method achieves the best mean accuracy.  As seen, when having a similarly smaller model complexity, our method beats PSTNet significantly. Besides,
our method built on ResNet152 achieved superior performance for most categories, e.g., the second best performance on ``Person'', ``Bike'', ``Curve'', ``'Bump' in IoU.
Most importantly,  the proposed method gained 0.4\% and 1.6\% improvements in mAcc and mIoU, respectively, against the current state-of-the-art MFTNet. 
The quantitative results verify that our method can extract better complementary cross-modality features.

% Figure \red{6} 
Figure~\ref{qua_mfnet} exhibits the qualitative results under different lighting conditions. 
In general, we find that our method has the following advantages. First, our method demonstrates better results than existing approaches for both night and daytime conditions. It shows slightly better performance for daytime images and more superior results for nighttime images. 
Second, our method can capture the tiny objects both in RGB and thermal images more effectively, such as the pedestrian in the 3rd column and the bump on the road in the 5th column. 
These advantages validate the effectiveness of our strategy for multi-modality feature fusion.

 Fig.~\ref{Attention Map} shows attention maps of three examples generated by our method and several baseline methods. It is shown that our method generates better attention than other approaches.

\paragraph{Results on PST900}
We then conduct experiments on PST900 dataset. We compare our method with Efficient FCN \cite{EfficientFCN}, CCNet \cite{CCNet}, ACNet \cite{Acnet}, SA-Gate \cite{SAGate}, RTFNet \cite{RTFNet}, PSTNet \cite{pst900}, and GMNet \cite{GMNet}. 
The quantitative results are given in Table~\ref{res_pst}.

It can be clearly seen that our method achieves the best results. It outperforms all previous methods, achieving 91.42 in mAcc and 85.56 in mIoU. Besides, it outperforms the state-of-the-art GMNet \cite{GMNet} by 1.81\% in mACC and 1.44\% in mIoU, respectively.
% clearly standing out among all the state-of-the-art approaches. 
% We also visualize several predicted semantic maps in Fig.~\ref{fig_pst}.
% In Figure 8, our model performs well on the PST900 dataset.


\subsection{Object Detection}
\subsubsection{Dataset}
M$^{3}$FD dataset \cite{m3fd}contains a set of auto-driving scenarios. It has 4200 pairs of RGB-T images, including 33603 annotated labels in six classes, including ``People'', ``Car'', ``Bus'', ``Motorcycle'', ``Truck'', and ``Lamp''. Moreover, the dataset was split into ``Daytime'', ``Overcast'', ``Night'', and ``challenge'' scenarios according to the characteristics of the environments.

\subsubsection{Implementation Details and Evaluation Metric}
We build a network for object detection by integrating EAEF into YoloV5 \cite{mmyolo}.  We use the stochastic gradient descent (SGD)\cite{SGD} optimization solver for training. The initial learning rate is set to 0.01, Momentum and weight decay are set to 0.9 and 0.0005, respectively. The batch size is set to 32, and we apply ExponentialLR to gradually decrease the learning rate. The loss function has an IoULoss \cite{Iouloss} term and a CrossEntropy \cite{celoss} term. These two loss terms are weighed with a scalar of 0.3 and 0.7, respectively. For evaluation, we take the mAP@0.5 metric as TarDAL \cite{m3fd}. 

\begin{table}[!t]
  \centering
  \caption{Quantitative results on the M3FD object detection dataset. }
    \begin{tabular}{cccccc}
    \toprule
    Method & Day   & Overcast & Night & Challenge & {mAP@0.5} \\
    \midrule
   RGB & 0.759 & 0.729 & 0.863 & 0.815 & 0.772 \\
   Thermal & 0.717 & 0.727 & 0.852 & \textbf{0.991} & 0.753 \\
   Yolov5\cite{Yolo} & 0.748  & 0.732 & 0.873 &0.867   & 0.763 \\
   U2F\cite{u2f} & 0.738 & 0.731 & 0.868 & 0.976 & 0.775\\
    TarDAL\cite{m3fd} & 0.745 & 0.741 & 0.893 & 0.983 & 0.778 \\
    Ours & \textbf{0.783} & \textbf{0.786} & \textbf{0.895} & 0.979 & \textbf{0.801} \\
    \bottomrule
    \end{tabular}%
  \label{res_m3fd}%
\end{table}%

\subsubsection{Results}

The experimental results are shown in Table~\ref{res_m3fd}. As seen, the method only using thermal data shows the worst performance.  Nevertheless, for ``Challenge'' scenarios,  it attains better performance than using RGB. Both TarDAL and our method obtain better accuracy compared to single modality data based methods.
It is also observed that our method outperforms the other approaches by a good margin. 
We obtain 0.801 mAP, outperforming TarDAL by 2.3\%.  

\subsection{Salient Objection Detection}

\subsubsection{Dataset}

We evaluate our method on VT821 \cite{MTMRVT821}, VT1000 \cite{SGDLVT1000}, and VT5000 \cite{ADFVT5000}, consisting of 821, 1000, and 5000 registered pairs of RGB-thermal images, respectively. 
Specifically, we utilize 2500 pairs from VT5000 for training and the remaining images, including those from the VT821 and VT1000 datasets, for testing.

\subsubsection{Implementation Details and Evaluation Metric}

We resize the image resolution to 224 × 224 and perform some data augmentations as LSNET  \cite{lSNet}, including the random flips, random rotations, and clipping. The framework is the same as that used in semantic segmentation tasks built on ResNet50.
We evaluate the performance using four metrics (S, adpE, adpF, MAE). The training hyperparameters, such as epoch, batch size, optimizer, and initial learning rate, are set to 20, 8, Adam optimization, and 0.001, respectively. 

\subsubsection{Results}
We provide a quantitative comparison against previous approaches, including MTMR \cite{MTMRVT821}, M3S-NIR \cite{M3SNIR}, SGDL \cite{SGDLVT1000}, FMCF \cite{FMCF}, ADF \cite{ADFNet}, MIDD \cite{MDDI}, ECFFNet \cite{ECFNet}, and LSNet \cite{lSNet}, CMDBIF-Net \cite{CMDBIF-Net}. As seen in Table~\ref{VT5000}, our method achieves the best performance on all metrics, e.g., outperforming the state-of-the-art CMDBIF-Net by 3.1$\%$ in MAE on VT5000.

%------------------------------------------------------------
%------------------------------------------------------------
\begin{table*}[ht]
	\centering
  \vspace{-3mm}
	\caption{Results of salient object detection on the VT821, VT1000, and VT5000 datasets.}
	\setlength{\tabcolsep}{1.3mm}{
		\begin{tabular}{lcccccccccccc}
			\toprule
			\multicolumn{1}{c}{\multirow{2}{*}{Method}} & \multicolumn{4}{c}{VT821}  & \multicolumn{4}{c}{VT1000} &
			\multicolumn{4}{c}{VT5000}\\
                \cmidrule(lr){2-5}\cmidrule(lr){6-9} \cmidrule(lr){10-13} 
			&\multicolumn{1}{c}{$S$$\uparrow$} & \multicolumn{1}{c}{$adpE$$\uparrow$} & \multicolumn{1}{c}{$adpF$$\uparrow$} & \multicolumn{1}{c}{MAE$\downarrow$}
			&\multicolumn{1}{c}{$S$$\uparrow$} & \multicolumn{1}{c}{$adpE$$\uparrow$} & \multicolumn{1}{c}{$adpF$$\uparrow$} & \multicolumn{1}{c}{MAE$\downarrow$}
			&\multicolumn{1}{c}{$S$$\uparrow$} & \multicolumn{1}{c}{$adpE$$\uparrow$} & \multicolumn{1}{c}{$adpF$$\uparrow$} & \multicolumn{1}{c}{MAE$\downarrow$} \\
			\hline
			MTMR\cite{MTMRVT821}
			& 0.725              & 0.815             & 0.662              & 0.109
			& 0.706              & 0.836              & 0.715             & 0.119
			& 0.680              & 0.795              & 0.595              & 0.114 \\
                M3S-NIR\cite{M3SNIR}
			& 0.723              & 0.859              & 0.734              & 0.140
			& 0.726             & 0.827             & 0.717             & 0.145
			& 0.652              & 0.780              & 0.575             & 0.168\\
                SGDL\cite{SGDLVT1000}
			& 0.765             & 0.847         & 0.731              & 0.085
			& 0.787              & 0.856             & 0.764             & 0.090
			& 0.750              & 0.824              & 0.672            & 0.089 \\
                FMCF\cite{FMCF}
			& 0.760              & 0.796             & 0.640            & 0.080
			& 0.873             & 0.899              & 0.823             & 0.037
			& 0.814             & 0.864              & 0.734             & 0.055 \\
                ADF\cite{ADFVT5000}
			& 0.810              & 0.842             & 0.717              & 0.077
			& 0.910             & 0.921             & 0.847              & 0.034
			& 0.864            & 0.891              & 0.778             & 0.048 \\ 
                MIDD\cite{MDDI}
			& 0.871              & 0.895              & 0.803              & 0.045
			& 0.915              & 0.933              & 0.880             & 0.027
			& 0.868              & 0.896              & 0.799             & 0.043 \\
                ECFFNet\cite{ECFNet}
			& 0.877              & 0.835              & 0.911              & 0.034
			& 0.924              & 0.919              & 0.959              & 0.021
			& 0.876              & 0.850              & 0.922              & 0.037 \\
                LSNet\cite{lSNet}
			& 0.877              & 0.911              & 0.827              & 0.033
			& 0.924              & 0.936              & 0.887              & 0.022
			& 0.876              & 0.916              & 0.827              & 0.036 \\
                CMDBIF-Net\cite{CMDBIF-Net}
			& 0.882             & 0.923             & 0.855              & 0.032
			& 0.927              & 0.952            &
              0.914              & 0.019
			& 0.886             & 0.933             & 0.868              & 0.032 \\
   %          \midrule
			% Ours(ResNet18)
			% & \textcolor[rgb]{ 1,  0,  0}{0.878} & \textcolor[rgb]{ 1,  0,  0}{0.917} & \textcolor[rgb]{ 1,  0,  0}{0.828} & \textcolor[rgb]{ 1,  0,  0}{0.032} 
   %  & \textcolor[rgb]{ 1,  0,  0}{0.918} & \textcolor[rgb]{ 1,  0,  0}{0.955}
			% & \textcolor[rgb]{ 1,  0,  0}{0.891} & \textcolor[rgb]{ 1,  0,  0}{0.020} & \textcolor[rgb]{ 1,  0,  0}{0.876}
			% & \textcolor[rgb]{ 1,  0,  0}{0.925} & \textcolor[rgb]{ 1,  0,  0}{0.832} & \textcolor[rgb]{ 1,  0,  0}{0.033} \\
            Ours
			& \textbf{0.885} & \textbf{0.927} & \textbf{0.846} & \textbf{0.031}
   
    & \textbf{0.926} & \textbf{0.964}
			& \textbf{0.905} & \textbf{0.017} & \textbf{0.885}
			& \textbf{0.934} & \textbf{0.853} & \textbf{0.031} \\
			\bottomrule
		\end{tabular}
	}
	\label{VT5000}
 \vspace{-3mm}
\end{table*}

\subsection{Crowd counting}

\subsubsection{Dataset}
RGBT-CC dataset \cite{iadm} has 2,030 RGB-T pairs captured in public scenarios. The images have a resolution of $640 \times\ 480$. A total of 138,389 pedestrians are marked with point annotations, and approximately 68 people are marked per image. The training, validation, and test set have 1545, 300, and 1200 RGB-T pairs, respectively.

\subsubsection{Implementation Details and Evaluation Metric}
 Following IADM \cite{iadm,CSCA}, we use the BL \cite{BL} network  built on the VGG16 \cite{vgg} as the backbone. We send the feature maps generated by the last EAEF module into an MLP decoder comprising two $1\times1$ convolutions to get the final prediction. The training strategy is also consistent with IADM \cite{iadm}. We use the Adam optimizer and set the learning rate to 0.00001. We evaluate the model every 10 epochs out of 300 epochs. The best model on the validation set will be used for evaluation. We measure with the root mean square error (RMSE) and the grid average mean absolute error (GAME) \cite{GAME}.

\begin{table}[!t]
  \centering
   \vspace{-3mm}
  \caption{Results on the RGBT-CC dataset. * denotes pertaining on the ImageNet.}
    \begin{tabular}
    % {ccccccc}
    {p{0.06\textwidth}<{\centering}p{0.055\textwidth}<{\centering}p{0.055\textwidth}<{\centering}p{0.055\textwidth}<{\centering}p{0.055\textwidth}<{\centering}p{0.055\textwidth}<{\centering}}
    \toprule
   Method & GAME(0)$\downarrow$ & GAME(1)$\downarrow$ & GAME(2)$\downarrow$ & GAME(3)$\downarrow$ & RMSE$\downarrow$
    \\
    \midrule
  UcNet\cite{UCNet} & 33.96 & 42.42 & 53.06 & 65.07 & 56.31 \\
    HDFNet\cite{HDFNet} & 22.36 & 27.79 & 33.68 & 42.48 & 33.93 \\
    BBSNet\cite{BBSNet} & 19.56 & 25.07 & 31.25 & 39.24 & 32.48 \\
   MVMS\cite{MVMS} & 19.97 & 25.1  & 31.02 & 38.91 & 33.97 \\
    % \midrule
    IADM\cite{iadm} & 15.61 & 19.95 & 24.69 & 32.89 & 28.18 \\
     CSCA\cite{CSCA} &14.32 &  18.91 & 23.81&  32.47 & 26.01\\
     TAFNet\cite{TAFNet}&  12.38 & 16.98 & 21.86 & 30.19 &  22.45\\
      Ours & 12.58 & 17.24 & 22.33 & 30.88 & 21.85\\
    \midrule
   DEFNet$^{*}$\cite{DEFNet} &11.90 &  16.08 & 20.19 & 27.27& 21.09 \\
   Ours$^{*}$ & \textbf{11.19} & \textbf{14.99} & \textbf{19.20} & \textbf{27.13} & \textbf{19.39} \\
    \bottomrule
    \end{tabular}%
  \label{res_rgbt}%
\end{table}%

% {\color{blue}
\subsubsection{Results}

 The results are given in Table~\ref{res_rgbt} where * denotes using pretrained VGG16 on the ImageNet. As shown, our method outperforms all baseline methods. Our method without and with pretraining outperforms TAFNet by 2.7\% and DEFNet by 8\% in RMSE, respectively. 
% }


%-------------------------------------------------------------------------
\subsection{Ablation Study}

We analyze the effectiveness of each component of our EAEF through additional experiments on the MFNet dataset. We establish a baseline by removing the AIB and ACB from the EAEF. The results are shown in Table~\ref{ablstudy}. We can observe that both AIB and ACB improved the performance of the baseline, and their combination, i.e., EAEF, gained the best performance.

\begin{table}[t]
  \centering
  \caption{Results of ablation study on the MFNet dataset.}

  \renewcommand\arraystretch{1.2}
    \begin{tabular}{cccc|cc}
    \toprule
    & Baseline & AIB & ACB & mAcc & mIoU \\
    \hline
    \multicolumn{2}{c}{$\surd$} &       &       & 71.7  & 56.5 \\
  \multicolumn{2}{c}{$\surd$} & $\surd$     &       & 72.5  & 57.1 \\
   \multicolumn{2}{c}{$\surd$} &       & $\surd$     & 74.3  & 57.7 \\
   \multicolumn{2}{c}{$\surd$} & $\surd$     & $\surd$     & \textbf{75.1} & \textbf{58.9} \\
    \bottomrule
    \end{tabular}%
   \label{ablstudy}
\end{table}%

\section{CONCLUSIONS}
In this paper, we studied the better fusion strategy of RGB images and thermal data for perception tasks.
We explicitly specify cases where i) both RGB and thermal data, ii) only one type of data, and iii) none of them can provide sufficiently useful features.
% Existing methods fuse RGB and thermal data/features by simply applying element-wise addition, concatenation, or a straightforward attention mechanism,  often yielding unsatisfactory performance. Such fusion approaches are insufficient and hard to analyze and diagnose. 
% We aim to take full advantage of multi-modality data and extract more complementary features. 
We proposed the explicit attention-enhanced fusion (EAEF) that enhances feature extraction and provides compensation for insufficient representations. We evaluated our method on different perception tasks, including semantic segmentation, object detection, salient detection, and crowd counting. As a result, we achieved state-of-the-art performance on all tasks, providing the robot community with a better fusion approach for RGB-thermal based perception tasks.

% Existing methods simply 

\bibliographystyle{IEEEtranS}
\bibliography{egbib}
% \end{thebibliography}

\ifCLASSOPTIONcaptionsoff
  \newpage
\fi



%\bibliographystyle{IEEEtran}


%\vfill


\end{document}